# A Hybrid ACO Algorithm for the Next Release Problem

He Jiang
School of Software
Dalian University of Technology
Dalian 116621, China
jianghe@dlut.edu.cn

Jingyuan Zhang
School of Software
Dalian University of Technology
Dalian 116621, China
zhangjy019@hotmail.com

Jifeng Xuan
School of Mathematical Sciences
Dalian University of Technology
Dalian 116024, China
xuan@mail.dlut.edu.cn

Zhilei Ren
School of Mathematical Sciences
Dalian University of Technology
Dalian 116024, China
ren@mail.dlut.edu.cn

Yan Hu
School of Software
Dalian University of Technology
Dalian 116621, China
huyan@dlut.edu.cn

*Abstract*—In this paper, we propose a Hybrid Ant Colony Optimization algorithm (HACO) for Next Release Problem (NRP). NRP, a NP-hard problem in requirement engineering, is to balance customer requests, resource constraints, and requirement dependencies by requirement selection. Inspired by the successes of Ant Colony Optimization algorithms (ACO) for solving NP-hard problems, we design our HACO to approximately solve NRP. Similar to traditional ACO algorithms, multiple artificial ants are employed to construct new solutions. During the solution construction phase, both pheromone trails and neighborhood information will be taken to determine the choices of every ant. In addition, a local search (first found hill climbing) is incorporated into HACO to improve the solution quality. Extensively wide experiments on typical NRP test instances show that HACO outperforms the existing algorithms (GRASP and simulated annealing) in terms of both solution quality and running time.

*Keywords-next release problem (NRP); ant colony optimization; local search; requirment engineering*

## I. INTRODUCTION

As a well-known problem arising in software requirement engineering, Next Release Problem (NRP) seeks to maximize the customer benefits from a set of interdependent requirements, under the constraint of a predefined budget bound. NRP was firstly formulated as an optimization problem in software evolution [1] by Bagnall in 2001 [2]. To optimize software requirements, NRP and its variants have attracted much attention from the research community in recent years [3-5], such as Component Selection and Prioritization [6], Fairness Analysis [7], Multi-Objective Next Release Problem (MONRP) [7, 8], and Release Planning [1, 9].

Since NRP has been proved as NP-hard, no exact algorithm exists to find optimal solutions in polynomial time unless P=NP [10]. Therefore, many heuristic algorithms have been proposed for NRP to obtain near optimal solutions in reasonable time, including greedy algorithms, hill climbing, simulated annealing (SA) [2, 6], and genetic algorithms [3, 8]. Among these algorithms, LMSA (a Simulated Annealing algorithm by Lundy and Mees) can work efficiently on some problem instances [2].

Ant Colony Optimization (ACO) is one of the new technologies in approximately solving NP-hard problems since 1991 [11]. With a colony of artificial ants, ACO can achieve good solutions for numerous hard discrete optimization problems. To improve the performance of ACO, some hybrid ACO algorithms (i.e. ACO with local search) have also been proposed for solving classical optimization problems, including Traveling Salesman Problem (TSP) [11], Quadratic Assignment Problem (QAP) [12], 3-Dimensional Assignment Problem (AP3) [13], etc.

Motivated by the great successes of ACO in tackling NP-hard problems, we propose a new algorithm named Hybrid ACO (HACO) for solving large NRP instances in this paper. HACO updates the solutions under the guideline of artificial ants and pheromone trails. At every iteration, HACO constructs the solutions and updates the pheromones for the next iteration. In contrast to traditional ACO algorithms, a first found hill climbing (FHC) operator is incorporated into HACO to improve the solution quality. This operator always updates solutions when the first better solution is found. Since it's a challenging and time consuming task to tune appropriate parameters for ACO algorithms, we employ CALIBRA procedure (an automatic tuning procedure by Adenso-Diaz and Laguna in 2006) [14] to effectively determine those interrelated parameter settings. Experimental results on typical NRP test instances show that our new algorithm outperforms those existing algorithms in both solution quality and running time.

The remainder of this paper is organized as follows. Section II gives out the related definitions of NRP. Section III introduces ACO and HACO for NRP. Section IV describes the

This work is partially supported by the Natural Science Foundation of China under Grant No. 60805024 and the National Research Foundation for the Doctoral Program of Higher Education of China under Grant No. 20070141020.

experimental results on NRP instances. Section V discusses some related work and Section VI concludes this paper.

## II. PRELIMINARIES

This section gives the definitions of NRP and some notations which will be used in the following part of this paper.

In a software system, let $R=\{r_1, r_2, \ldots, r_n\}$ denotes all the possible software requirements. Each requirement $r_i \in R$ is associated with cost $c_i \in Z^+$. A directed acyclic graph $G=(R,E)$ denotes the dependency of requirements, where $(r, r') \in E$ indicates that $r$ must be satisfied before $r'$. A set of requirements $parent(r)$ contains all requirements which must be satisfied before $r$.

Let $S=\{1, 2, \ldots, m\}$ denotes all the customers related to the requirements. Each customer $i$ is satisfied, if and only if a set of requirements $R_i \subseteq R$ is satisfied. A profit $w_i \in Z^+$ denotes the priority of customer $i$. To satisfy a customer, both the requirements of $R_i$ and the ones satisfied before them $parent(R_i) = \bigcup_{r \in R_i} parent(r)$ must be satisfied. Let $R_i' = R_i \cup parent(R_i)$ denotes all the requirements which have to be developed to satisfy customer $i$. Let $cost(R_i') = \sum_{r_j \in R_i'} c_j$ be the cost of satisfying customer $i$. The cost of a subset $S' \subseteq S$ is defined as $cost(S') = cost(\bigcup_{i \in S'} R_i')$ and the overall profit obtained is defined as $\omega(S') = \sum_{i \in S'} w_i$.

Based on above definitions, the goal of NRP is to find a subset $S' \subseteq S$ to maximize $\omega(S')$, subject to $cost(S') \leq B$, where $B$ is the predefined development budget bound [2]. Given a NRP instance (denoted as $NRP(S,R,W)$), a feasible solution is a subset $S' \subseteq S$ subject to $cost(S') \leq B$.

## III. A HYBRID ACO ALGORITHM

In this section, we present the HACO algorithm for solving NRP in detail. In contrast to traditional ACO algorithms, a local search operator called FHC is incorporated into HACO to improve the quality of solutions. We present the framework of this HACO in Part A and the local search operator in Part B, respectively.

### A. HACO

Ant Colony Optimization (ACO) is a meta-heuristic algorithm based on communication of a colony of simple agents (artificial ants), which are directed by artificial pheromone trails. The ants use pheromone trails and heuristic information to probabilistically construct solutions. In addition, the ants also use these pheromone trails and heuristic information during the algorithm's execution to reflect their search experiences [11, 12].

In this subsection, we propose HACO algorithm to solve NRP. The main framework of HACO is presented in Algorithm 1. After the initialization, a series of iterations are conducted to find solutions. At every iteration, HACO employs those ants to construct solutions by a probabilistic action choice rule and incorporates a local search operator $L$ to further improve those constructed solutions (see Step (2.1)). Then, HACO updates the pheromone trails for the next iteration, including the pheromone evaporation (see Step (2.2)) and deposition (see Step (2.3)). During the iterations, when a better solution is found, our best solution will be updated. Finally, the best solution will be returned.

---

**Algorithm 1:** HACO
**Input:** NRP($S,R,W$), local search operator $L$, iteration times $t$, ant number $h$
**Output:** Solution $S^*$
**Begin**
(1) $S^* = \varnothing$, $\omega(S^*) = 0$;
(2) for $i$=1 to $t$ do
  (2.1) for $k$=1 to $h$ do
       // *SolutionConstruction phase*
       ant $k$ construct solutions $S_k$;
       // *Local search phase*
       call $L$ to optimize $S_k$;
  //*Pheromone update phase*
  (2.2) evaporate pheromone on all customers, by (2);
  (2.3) for $k$=1 to $h$ do
       deposit pheromone on customers in $S_k$, by (4);
       if $\omega(S^*) < \omega(S_k)$ then $S^* = S_k$;
(3) return $S^*$;
**End**

---

More details related to HACO are discussed as follows:

*1) Pheromone trail and heuristic information:* The pheromone trails $\tau_i$ for NRP refers to the desirability of adding a customer $i$ to the current partial solution. We define $\tau_i = \theta w_i$, where $\theta$ is a parameter associated with customer profits $w_i$ to make sure that the initial pheromone value varies between 0 and 1. We also define the heuristic information $\eta_i = w_i/cost(R_i')$. Obviously, the higher profit and the lower requirements cost a customer has, the higher $\eta_i$ is.

*2) Solution Construction:* Initially, there're no customers in ants for NRP. At each construction step, ant $k$ applies a probabilistic action choice rule (random proportional rule [11]) to decide which customer to add next.

In particular, the probability with which ant $k$ chooses customer $i$ is given by:

$$p_i^k = \frac{[\tau_i]^\alpha [\eta_i]^\beta}{\sum_{l \in N^k} [\tau_l]^\alpha [\eta_l]^\beta}, \text{ if } i \in N^k. \quad (1)$$

where $\alpha$ and $\beta$ are two parameters which determine the relative influences of the pheromone trail and the heuristic information, and $N^k$ is the set of customers that ant $k$ has not chosen yet.

The pseudo-code for the ant solution construction is given in Algorithm 2. This procedure works as follows. Firstly, there is no customer in ant $k$ (see Step (1)). Secondly, the next customer is chosen probabilistically by the roulette wheel selection procedure of evolutionary computation, proposed by Goldberg in 1989 [15]. The roulette wheel is divided into slices proportional to weights of customers that ant $k$ has not chosen yet (see Step (2.3.2)). Finally, the wheel is rotated and the next customer $i$ chosen by ant $k$ is the one which the marker points to (see Step (2.3.3)).

---

**Algorithm 2:** SolutionConstruction
**Input:** NRP($S,R,W$), budget $B$, ant $k$
**Output:** Solution $S^*$
**Begin**
(1) $S^* = \emptyset$, $\omega(S^*) = 0$;
(2) while (cost($S^*$) ≤ $B$)
  (2.1) $sumProb = 0$;
  (2.2) let $r$ be a value randomly generated in [0, 1];
  (2.3) for each customer $i$ do
    (2.3.1) if ($i$ has been added to $S^*$)
        $p_i^k = 0.0$;
    (2.3.2) else
        $p_i^k$ is calculated by (1);
        $sumProb\mathrel{+}= p_i^k$;
    (2.3.3) if ($sumProb \geq r$)
        add customer $i$ to $S^*$, break;
(3) return $S^*$;
**End**

---

*3) Update of pheromone trails:* The pheromone trails are updated after all the ants have constructed their solutions. Firstly, we decrease the pheromone value of every customer by a constant factor, and then add pheromone to those customers that the ants have chosen in their solutions.

Pheromone evaporation is implemented by

$$\tau_i \leftarrow (1 - \rho)\, \tau_i. \qquad (2)$$

where $0 < \rho \leq 1$ is the pheromone evaporation rate, which enables the algorithm to forget bad decisions previously taken and have more opportunities to choose other customers.

All ants deposit pheromone on the customers which they have chosen in their solutions after evaporation:

$$\tau_i \leftarrow \tau_i + \sum_{k=1}^{h} \Delta \tau_i^k. \qquad (3)$$

where $\Delta \tau_i^k$ is the amount of pheromone that ant $k$ deposits on those chosen customers. It is defined as follows:

$$\Delta \tau_i^k = \begin{cases} \gamma W^k, & i \text{ is selected} \\ 0, & \text{otherwise} \end{cases} \qquad (4)$$

where $W^k$, the quality of the solution $S_k$ built by ant $k$, is computed as the sum of customer profits in $S_k$. The parameter $\gamma$ is used to tune $\Delta \tau_i^k$.

### B. FHC

To enhance the performance of HACO, we incorporate a local search operator $L$ named FHC into HACO. In this section, we introduce the framework of FHC (see Algorithm 3).

---

**Algorithm 3:** FHC
**Input:** NRP($S,R,W$), budget $B$, iteration times $t$
**Output:** Solution $S^*$
**Begin**
(1) $S^* = \emptyset$, $\omega(S^*) = 0$;
(2) for $i$=1 to $t$ do
  (2.1) let $S_0$ be a random feasible solution;
  (2.2) flag = true;
  (2.3) while (flag = true) do
    (2.3.1) randomly choose $j \in S \setminus S_0$;
    (2.3.2) let $S_1 = S_0 \cup \{j\}$;
    // when $S_1$ is feasible
    (2.3.3) if $cost(S_1) < B$ then $S_0 = S_1$;
        else // swap a customer in $S_0$ with $j$
    (2.3.3.1) flag = false;
    (2.3.3.2) $S_1 = S_0$;
    (2.3.3.3) for every customer $l \in S_0$ do
        if $cost(S_0 \cup \{j\} \setminus \{l\}) < B$ and
$\omega(S_0 \cup \{j\} \setminus \{l\}) > \omega(S_1)$
        then $S_1 = S_0 \cup \{j\} \setminus \{l\}$, flag = true;
    (2.3.3.4) $S_0 = S_1$;
  (2.4) if $\omega(S_0) > \omega(S^*)$ then $S^* = S_0$;
(3) return $S^*$;
**End**

---

The hill climbing algorithm is a classic local search technology, which can find local optimal solutions in the solution neighborhood. FHC always updates the solution when a first local optimal solution is found.

FHC mainly consists of a series of iterations. At every iteration, a feasible solution will be randomly generated as the current solution (see Step (2.1)). After that, this current solution $S_0$ will be further improved by local search as follows (see Step (2.3)). Firstly, an unselected customer $j \in S \setminus S_0$ will be arbitrarily chosen out (see Step (2.3.1)). A new solution can be achieved by adding customer $j$ to $S_0$ (see Step (2.3.2)). If such action will result in a new feasible solution, then the current solution $S_0$ will be updated with the new generated solution and the local search process is further conducted to improve it (see Step (2.3.3)). Otherwise, we try to replace a customer $l \in S_0$ with $j$ such that the resulting solution is feasible and its profit is maximized. If succeeded, the current solution will be updated with the resulting solution and the local search is further conducted to improve it (see Step (2.3.3.1)-(2.3.3.4)). After every iteration, when a better solution is obtained, the best solution will be updated (see Step (2.4)). Finally, the best solution is returned (see Step (3)).

TABLE I. THE DETAILS OF INSTANCE GENERATION

| Instance | NRP-1 | NRP-2 | NRP-3 | NRP-4 | NRP-5 |
|---|---|---|---|---|---|
| Number | 20/40/80 | 20/40/80/160/320 | 250/500/750 | 250/500/750/1000/750 | 500/500/500 |
| Cost | 1~5/2~8/5~10 | 1~5/2~7/3~9/4~10/5~15 | 1~5/2~8/5~10 | 1~5/2~7/3~9/4~10/5~15 | 1~3/2/3~5 |
| Max. | 8/2/0 | 8/6/4/2/0 | 8/2/0 | 8/6/4/2/0 | 4/4/0 |
| Customer | 100 | 500 | 500 | 750 | 1000 |
| Request | 1~5 | 1~5 | 1~5 | 1~5 | 1 |
| Profit | 1~30 | 1~30 | 1~30 | 1~30 | 1~30 |

## IV. EXPERIMENTAL RESULTS

In this section, we describe the experimental results to evaluate the performance of HACO and existing algorithms.

### A. Instances generation

Since the customer requirements are usually private data of a company, no public NRP instance is available. We generate NRP test instances by following the methods in the classic paper of NRP [2]. These instances consist of five randomly generated problems, each of which contains some multi-level requirements. Each requirement in a level is with the constraint of numbers, dependency, and cost. For each problem, the predefined development budget bound is defined as 30%, 50%, and 70% of the total cost of requirements. But the ranges of customer profits have not been mentioned before. We assume each customer profit is selected randomly from 1 to 30. Table I gives the details of instance generation. The 6 rows of this table show the instance names, number of requirement in each level, cost of requirements, the maximum number of requirement dependency, number of customers, request of each customer, and the customer profits.

### B. Parameters for HACO

The performance of ACO algorithms usually depends on the configuration of five parameters. In this section, we use CALIBRA procedure [14], proposed by Adenso-Diaz and Laguna in 2006, to determine the interrelated HACO parameter settings that improve the algorithm defaults.

CALIBRA is an automated tool for finding performance-optimizing parameter settings, which liberates algorithm designers from the tedious task of manually searching the parameter space. It uses Taguchi's fractional factorial experimental designs coupled with local search to finely tune algorithm parameters with a wide range of possible values. The benefit of using CALIBRA is evident in situations where the algorithm being fine-tuned has parameters whose values have significant impact on performance.

When linking the online supplement and using the current version of CALIBRA[1], 5 instances are selected randomly for testing, one for each group size, to ensure proper representation of problem instances from all sizes. We pick lower and upper bounds for parameters and discretize the intervals uniformly. For example, the continuous pheromone influence parameter $\alpha$ in the range 0.1 to 4.0 with an accuracy of one decimal place is internally handled as an integer variable with a lower bound of 1 and an upper bound of 40. The other four parameters are handled in the same way. Table II shows the parameters, the original range considered before discretization and the values found by CALIBRA. With these parameter values, HACO can achieve good solutions on all problem instances.

TABLE II. HACO PARAMETER RANGES CONSIDERED FOR TUNING AND RESULTS OF RUNNING CALIBRA

| Parameters | Ranges | Values by CALIBRA |
|---|---|---|
| $\alpha$ | [0.1, 4.0] | 1.1 |
| $\beta$ | [0.1, 6.0] | 1.5 |
| $\gamma$ | [0.010, 0.050] | 0.020 |
| $\rho$ | [0.01, 0.99] | 0.13 |
| $h$ | [1, 30] | 10 |

### C. Results and analysis

In this section, we show the experimental results of heuristic algorithms, including GRASP, SA [2], FHC, ACO, and HACO, where ACO is referred as the HACO version using no local search. All the algorithms are implemented in C++ and run on a personal computer with Microsoft Windows XP, Intel Core 2.53GHz CPU and 4GB memory. For those heuristic algorithms, we set their parameters as follows. Both in GRASP and in FHC, each solution quality is the best found after 100 re-starting times. For GRASP, the construction phase is implemented according to [2] and the length of Restricted Candidate List (RCL) is set to 10. The local search phase is similar to FHC. For SA algorithm, the Lundy and Mees cooling schedule in [2] is conducted with temperature control parameter to be $10^{-8}$. In ACO and HACO, we set its five parameters to 1.1, 1.5, 0.020, 0.13, and 10 as suggested by CALIBRA. For the HACO, 10 iterations are executed and 10 ants apply local search at each iteration for comparison with FHC within similar running time.

Table III illustrates ACO and HACO outperform other algorithms in terms of solution quality and running time. The first column shows the instance names, and the following 5 columns show the 5 algorithms in the experiments. The 2 sub-columns of each algorithm present the solution quality and running time, respectively.

---

[1] http://or.pubs.informs.org/org/Pages.collect.html

TABLE III. EXPERIMENTAL RESULTS OF FIVE ALGORITHMS ON NRP INSTANCES

| Instance | GRASP | | SA | | FHC | | ACO | | HACO | |
|---|---|---|---|---|---|---|---|---|---|---|
| | *solution* | *time(s)* | *solution* | *time(s)* | *solution* | *time(s)* | *solution* | *time(s)* | *solution* | *time(s)* |
| NRP-1-0.3 | 810 | 1.406 | 827 | 14.453 | 817 | 0.968 | 829 | 1.703 | 835 | 0.734 |
| NRP-1-0.5 | 1364 | 1.281 | 1397 | 27.312 | 1353 | 1.281 | 1416 | 1.671 | 1418 | 0.89 |
| NRP-1-0.7 | 1945 | 1.359 | 1960 | 32.921 | 1939 | 0.812 | 1963 | 1.656 | 1968 | 0.859 |
| NRP-2-0.3 | 4084 | 144.328 | 3902 | 401.031 | 3802 | 334.062 | 4139 | 42.593 | 4262 | 187.218 |
| NRP-2-0.5 | 6616 | 333.187 | 6629 | 655.687 | 6352 | 401.64 | 6712 | 43.125 | 6786 | 337.484 |
| NRP-2-0.7 | 9284 | 351.437 | 9602 | 801.859 | 9436 | 349.937 | 9628 | 42.5 | 9732 | 274.593 |
| NRP-3-0.3 | 6061 | 140.25 | 5556 | 820.812 | 5423 | 678.39 | 6063 | 48.984 | 6067 | 397.531 |
| NRP-3-0.5 | 9131 | 300.14 | 8836 | 940.687 | 8647 | 666.921 | 9136 | 48.812 | 9196 | 464.312 |
| NRP-3-0.7 | 11723 | 102.078 | 11648 | 352.312 | 11572 | 191.25 | 11726 | 47.187 | 11728 | 79.812 |
| NRP-4-0.3 | 9161 | 1329.59 | 8498 | 5763.55 | 7935 | 4729.84 | 9165 | 137.156 | 9183 | 2810.8 |
| NRP-4-0.5 | 13731 | 1220.11 | 13279 | 8827.31 | 12822 | 12208.4 | 13751 | 136.968 | 13810 | 3059.88 |
| NRP-4-0.7 | 17999 | 273.312 | 17932 | 1809.71 | 17771 | 1433.06 | 18022 | 131.156 | 18029 | 424.234 |
| NRP-5-0.3 | 14940 | 465.14 | 14808 | 2535.55 | 13322 | 506.296 | 15592 | 158.515 | 15616 | 282.76 |
| NRP-5-0.5 | 20030 | 686.718 | 19991 | 2682.19 | 19608 | 1461.38 | 20784 | 156.531 | 20946 | 1036.47 |
| NRP-5-0.7 | 24508 | 138.203 | 24439 | 375.656 | 24410 | 245.781 | 24568 | 155.578 | 24570 | 163.312 |

On the small instances as NPR-1, the running time of ACO is nearly the same as the existing ones, but on large NRP instances, ACO algorithm can get better solutions in much less computing time than other algorithms. However, HACO can achieve better solutions than ACO and need less running time to obtain a local optimal solution than FHC.

Fig. 1 shows the solution quality comparison of FHC, ACO and HACO algorithms in a more intuitively way. ACO and HACO prove to be much more sufficient than FHC on all instances, especially on large ones like NRP-4 and NRP-5. HACO can obtain the best solution quality on all instances among these 3 algorithms and can be implemented as a general approach for NRP problems.

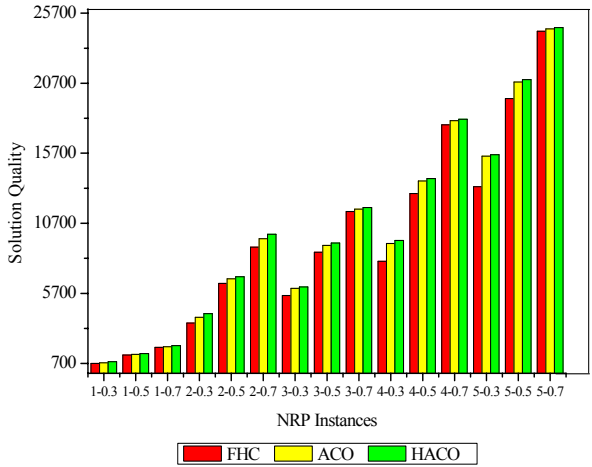

Figure 1. Solution qualities obtained by FHC, ACO and HACO algorithms on all instances.

Fig. 2 illustrates the running time of FHC, ACO and HACO algorithms. On all the NRP instances, ACO uses the shortest computing time. The running time of FHC on larger instances like NRP-4 is much longer than that of HACO. It implies that when we incorporate local search operator FHC into HACO, those high-quality solutions constructed by ACO can reduce the time of searching for local optima to some extent.

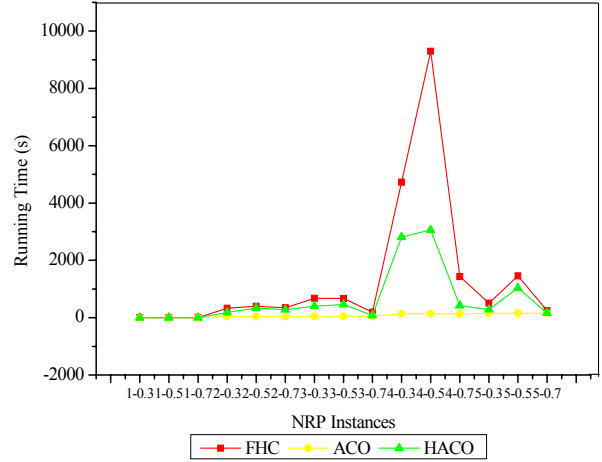

Figure 2. An illustration of running time by FHC, ACO and HACO algorithms on all instances.

## V. RELATED WORK

There already existed several heuristic approaches for the NRP problem. Among those algorithms, the hill climbing (HC) and simulated annealing (SA) were implemented in [2, 6]. When using greedy algorithm in [2] as the construction phase, GRASP can find better solutions than SA on some larger instances within only modest amounts of computing time (as shown in Table III in Section IV). Recently, an approximate backbone based multilevel algorithm (ABMA) was proposed in [17] to solve large scale NRP instances.

ACO is a kind of meta-heuristic approach and Ant System (AS) [18] was the first ACO algorithm, using Traveling Salesman Problem (TSP) as an example application. Then a number of direct extensions of AS were provided [19-21], being different in the way the pheromone updated, as well as the management details of the pheromone trails.

When using ACO algorithms, automatic tuning procedures could effectively find better parameter settings. There already existed several approaches for automatic algorithm parameters tuning. Boyan and Moore [22] proposed a tuning algorithm based on machine learning techniques. But there was no empirical analysis of this algorithm when applied to parameters that have wide range of possible values. Audet and Orban [23] introduced a mesh adaptive direct search that used surrogate models for algorithmic tuning. Nevertheless, this approach has never been used for tuning local search algorithms. Adenso-Diaz and Laguna [14] designed an algorithm called CALIBRA for fine tuning algorithms. It can narrow down choices for parameter values when facing a wide range of possible values and search for the best possible parameter settings.

## VI. Conclusion and Future Work

In this paper, we propose HACO for NRP, which incorporates FHC into ACO to solve NRP. We compare the performance of standard ACO and HACO with existing heuristic algorithms on some typical NRP instances. The experimental results illustrate that ACO outperforms the existing algorithms in solution quality and running time. When coupled with FHC, HACO can achieve better solution quality than ACO.

In future work, we will investigate how to further improve the performance of HACO and extend HACO for multi-objective NRP.

## Acknowledgment

We would like to thank Prof. Belarmino Adenso-Diaz and Prof. Manuel Laguna for providing the latest online version of CALIBRA. Prof. Adenso-Diaz is with Computer Engineering School and Industrial Engineering School, University of Oviedo and Prof. Laguna is with Leeds School of Business, University of Colorado.